\documentclass[runningheads]{llncs}

\usepackage{eccv}

\usepackage{eccvabbrv}

\usepackage{graphicx}
\usepackage{booktabs}

\usepackage[subrefformat=parens]{subcaption}
\usepackage[accsupp]{axessibility}  % Improves PDF readability for those with disabilities.

\usepackage{hyperref}

\usepackage{orcidlink}
\usepackage{booktabs}
\usepackage{bm}

\usepackage{tikz}
\usepackage[option]{colortbl}

\DeclareMathOperator*{\argmax}{argmax}

\usepackage{mathtools}

\begin{document}

\title{Revisiting Relevance Feedback for \\ CLIP-based Interactive Image Retrieval} 


\author{Ryoya Nara\inst{1}\and
Yu-Chieh Lin\inst{2} \and
Yuji Nozawa\inst{2} \and
Youyang Ng\inst{2} \and
Goh Itoh\inst{2} \and \\
Osamu Torii\inst{2} \and
Yusuke Matsui\inst{1}
}

\authorrunning{R.~Nara et al.}

\institute{The University of Tokyo \and
Kioxia Corporation
}

\maketitle

\begin{abstract}
Many image retrieval studies use metric learning to train an image encoder. However, metric learning cannot handle differences in users' preferences, and requires data to train an image encoder. To overcome these limitations, we revisit relevance feedback, a classic technique for interactive retrieval systems, and propose an interactive CLIP-based image retrieval system with relevance feedback. Our retrieval system first executes the retrieval, collects each user's unique preferences through binary feedback, and returns images the user prefers. Even when users have various preferences, our retrieval system learns each user's preference through the feedback and adapts to the preference.
Moreover, our retrieval system leverages CLIP's zero-shot transferability and achieves high accuracy without training. 
We empirically show that our retrieval system competes well with state-of-the-art metric learning in category-based image retrieval, despite not training image encoders specifically for each dataset. Furthermore, we set up two additional experimental settings where users have various preferences: one-label-based image retrieval and conditioned image retrieval. In both cases, our retrieval system effectively adapts to each user's preferences, resulting in improved accuracy compared to image retrieval without feedback. Overall, our work highlights the potential benefits of integrating CLIP with classic relevance feedback techniques to enhance image retrieval.

\end{abstract}

\section{Introduction}
\label{sec:intro}

With the rapid growth of available images in recent years, image retrieval systems have become ubiquitous, particularly in e-commerce and internet search applications. Image retrieval systems receive a query image from a user, retrieve similar images, and return them to the user.

Many content-based image retrieval studies utilize metric learning~\cite{proxy-nca,multi-similarity,triplet-deep}.  However, in many cases, metric learning is insufficient because users may have various preferences for the returned images. For example, one may search for dog images by querying an image of a dog running in a park, and another may search for park images with the same query image. Simply mapping images into image features might not accommodate those varying preferences. Furthermore, metric learning requires abundant annotated data to train an image encoder.

To overcome these limitations of metric learning-based image retrieval methods, we revisit the classic relevance feedback~\cite{relevance-feedback-revisited} technique by incorporating the Contrastive Language-Image Pre-training (CLIP) model~\cite{CLIP}. We propose an image retrieval system that encodes images with a CLIP image encoder and employs relevance feedback to gather information about each user's preferences. The overall retrieval process is as follows:
\begin{enumerate}
    \item Our retrieval system receives a query image from a user, retrieves images from a database, and returns them to the user.
    \item The user reviews the returned images and provides binary feedback on each image, indicating whether the user prefers it. 
    \item The system learns the user's preference and returns images the user prefers.
\end{enumerate}

While our retrieval system requires users to provide feedback on the returned samples, our system can adapt to each user's unique preference through the users' feedback, even when users have various preferences. Whenever a user provides a query image, our system learns the user's preference in real time and returns relevant images that align with the preference. Moreover, our retrieval system does not necessitate training. We can achieve high retrieval accuracy by leveraging CLIP's zero-shot transferability and incorporating the user's feedback. Additionally, because our system does not rely on injecting preferences through text, it can handle diverse and complex queries while avoiding the limitations of multimodal retrieval methods~\cite{tirg,ComposeAE,genecis}.

Carefully following the classic test and control method~\cite{relevance-feedback-evaluation-survey,chang1971evaluation} for relevance feedback evaluation, we empirically show that our retrieval system achieves high accuracy in various settings, which are designed to represent different user behaviors. We evaluate our retrieval system by automatically generating users' feedback from dataset labels to simulate users' feedback. In category-based image retrieval settings, our retrieval system achieves competitive accuracy with state-of-the-art metric learning methods. Additionally, we set up two experimental settings where users have various preferences: one-label-based image retrieval and conditioned image retrieval. In these two experimental settings, our retrieval system achieves higher accuracy than image retrieval without relevance feedback. 

In this work, we aim to cast a spotlight on relevance feedback again. Despite the significant advancements in deep learning techniques for image retrieval, relevance feedback has received less attention in recent years. However, our findings demonstrate that incorporating relevance feedback into modern deep learning-based image retrieval systems, such as those based on CLIP, can significantly improve their accuracy. While this paper primarily focuses on incorporating relevance feedback into CLIP, our proposed method can also be applied to other pre-trained image encoders.

To sum up, our contributions are as follows:
\begin{itemize}
    \item We propose CLIP-based interactive image retrieval with relevance feedback. 
    \item We propose an evaluation method of image retrieval systems with relevance feedback to simulate users' feedback.
   \item With a realistic feedback size, our system achieves competitive results with supervised metric learning methods in category-based image retrieval, despite not training image encoders specifically for each dataset.
    \item We set up two experimental settings where users have various preferences. In both settings, our retrieval system improves accuracy from image retrieval without user feedback.
    \item With a realistic feedback size, our retrieval system achieves competitive results with state-of-the-art multimodal retrieval in conditioned image retrieval settings, despite not exploiting textual information.
\end{itemize}

\section{Related Work}

\subsection{Metric Learning}

Metric learning~\cite{hist,triplet} is a major approach to training an image encoder for image retrieval systems. Metric learning trains an image encoder to map an image into an image feature so that semantically similar images are close together and dissimilar images are apart. Metric learning are utilized in image retrieval~\cite{proxy-anchor}, personal re-identification~\cite{beyond-personal-reidentification}, face recognition~\cite{Magface}, landmark retrieval~\cite{gldv2}, and few-shot learning~\cite{few-show-feature-map}. 

\subsection{Interactive Image Retrieval}

Interactive image retrieval systems enable users to inject their preferences into the retrieval system and obtain samples they prefer. Before the advent of deep learning, relevance feedback~\cite{relevance-feedbacl-powerful-tool,relevance-feedback-revisited} had been a popular technique for learning each user's preference. Recently, multimodal retrieval~\cite{compose-ae,pic2word} has become mainstream of interactive image retrieval.

Relevance feedback~\cite{click-feedback-retrieval,deep-relevance-feedback} is a classic technique for interactive retrieval systems. Each user provides feedback for the returned samples, and the retrieval system receives the user's feedback and executes retrieval again to meet their preferences better. In addition to one-time binary feedback, Ahmed~\cite{relevance-feedback-medical} used scalers for multi-level feedback, and Wu \etal~\cite{multilevel-relevance-feedback} studied retrieval systems where users provide feedback multiple times.
Our work revisits relevance feedback. We focus on cases where users provide binary feedback just once.

Multimodal retrieval~\cite{genecis,pic2word} enables users to inject their preferences through text. For example, along with a query image of a long red dress, a user injects a text query of 
 ``I want something short and yellow.'' In response, the image retrieval system returns an image of a short yellow dress. This paper compares our relevance-feedback-based retrieval with multimodal retrieval in \cref{subsec:conditioned}.

\subsection{Contrastive Language-Image Pre-training (CLIP) }

Contrastive language-image pre-training (CLIP)~\cite{CLIP} is a pre-trained vision-language model trained on large-scale image-text pairs. CLIP consists of an image encoder that embeds images into a feature space and a text encoder that embeds strings into the same feature space. CLIP models achieve impressive results on various downstream tasks~\cite{clipcap,sb-zsir}, and many studies leverage CLIP as powerful feature extractors.

\section{Approach}

First, we explain the overall retrieval pipeline in \cref{subsec:approach-problem-settings}. Next, we describe how we develop a retrieval algorithm that adapts to each user's preference in \cref{subsec:approach-proposed-method}. Finally, we explain how to accurately assess whether our retrieval system aligns with user preferences in \cref{subsec:approach-evaluation}.

\subsection{Retrieval Pipeline}\label{subsec:approach-problem-settings}

We execute retrieval twice for each query image $q \in \mathcal{I}$. Here, $\mathcal{I}$ is a space representing all possible images. We consider two databases: $\mathcal{X}_1, \mathcal{X}_2 \subset \mathcal{I}$. First, our retrieval system executes retrieval for database images $\mathcal{X}_1$. Next, it updates the retrieval algorithm and executes retrieval for database images $\mathcal{X}_2$. In actual application, the database images in both retrievals are the same: $\mathcal{X}_1 = \mathcal{X}_2$, i.e., we have just one single database $\mathcal{X}$ and execute retrieval for $\mathcal{X}$ twice. We introduce the two databases here for a fair comparison, as we will explain in \cref{subsec:approach-evaluation}.  

In advance, we prepare database features. We use visual encoder $\mathbf{\phi}: \mathcal{I} \to \mathbb{R}^D$ and get database features for the first retrieval $\mathcal{V}_1\coloneq \ \{\mathbf{\phi}(x) \mid x \in \mathcal{X}_1 \} \subset \mathbb{R}^D$, and database features for the second retrieval $\mathcal{V}_2 \coloneq \{ \mathbf{\phi}(x) \mid x \in \mathcal{X}_2 \} \subset \mathbb{R}^D$. 

When a user provides a query image $q$, our retrieval system encodes $q$ into $D$-dimensional feature as $\mathbf{u} \coloneq \mathbf{\phi} (q)  \in \mathbb{R}^D$. Our retrieval system executes retrieval for $\mathcal{V}_1$ to obtain similar features to $\mathbf{u}$. Let us denote the top-$M$ similar samples to $\mathbf{u}$ from $\mathcal{V}_1$ as $\mathcal{W}_1 \subseteq \mathcal{V}_1$. We write this K-NN (Nearest Neighbor) retrieval operation as a function form in \cref{eq:W_1}:
\begin{equation}\label{eq:W_1}
    \mathcal{W}_1 =
    \psi(\mathbf{u}, M, \mathcal{V}_\mathrm{1}) \coloneq
    M\text{-}\argmax_{\mathbf{w} \in \mathcal{V}_1} \frac{\mathbf{u}^\top \mathbf{w}}{\Vert \mathbf{u} \Vert_2 \Vert \mathbf{w} \Vert_2}.
\end{equation}
Note that $|\mathcal{W}_1|=M$. Our retrieval system returns $\mathcal{W}_1$ to the user.

Next, the user provides binary feedback to each sample in $\mathcal{W}_1$. We formulate the feedback as \cref{eq:feedback-form}:
\begin{equation}\label{eq:feedback-form}
    \mathcal{F} = \{ (\mathbf{w}, b) \mid \mathbf{w} \in \mathcal{W}_1 \}.
\end{equation}
Here, $b \in \{0, 1\}$ is the user's feedback. If the user prefers a returned sample $\mathbf{w}$, they provide $b=1$ to $\mathbf{w}$. Otherwise, they provide $b=0$. We can regard $\mathcal{F}$ as a labeled dataset where each sample $\mathbf{w}$ has a binary label $b$.

With $\mathcal{F}$, we update the retrieval algorithm $\psi$ to $\tilde{\psi}$. Next, our retrieval system executes retrieval for $\mathcal{V}_2$ and retrieves top-$K$ sample $\mathcal{W}_2 \subset \mathcal{V}_2$:
\begin{equation}\label{eq:updated-retrieval-definition}
    \mathcal{W}_2 = \tilde{\psi}(\mathbf{u}, K, \mathcal{V}_2).
\end{equation}
$\mathcal{W}_2$ represents the final returned samples. We aim to update the retrieval algorithm so that each returned sample in $\mathcal{W}_2$ is preferable for the user. Moreover, we aim to choose the visual encoder $\mathbf{\phi}$ to achieve high retrieval performance in various user preferences and datasets.

\subsection{Proposed Method}\label{subsec:approach-proposed-method}

\begin{figure}[tb]
    \centering
    \includegraphics[width=0.9\linewidth]{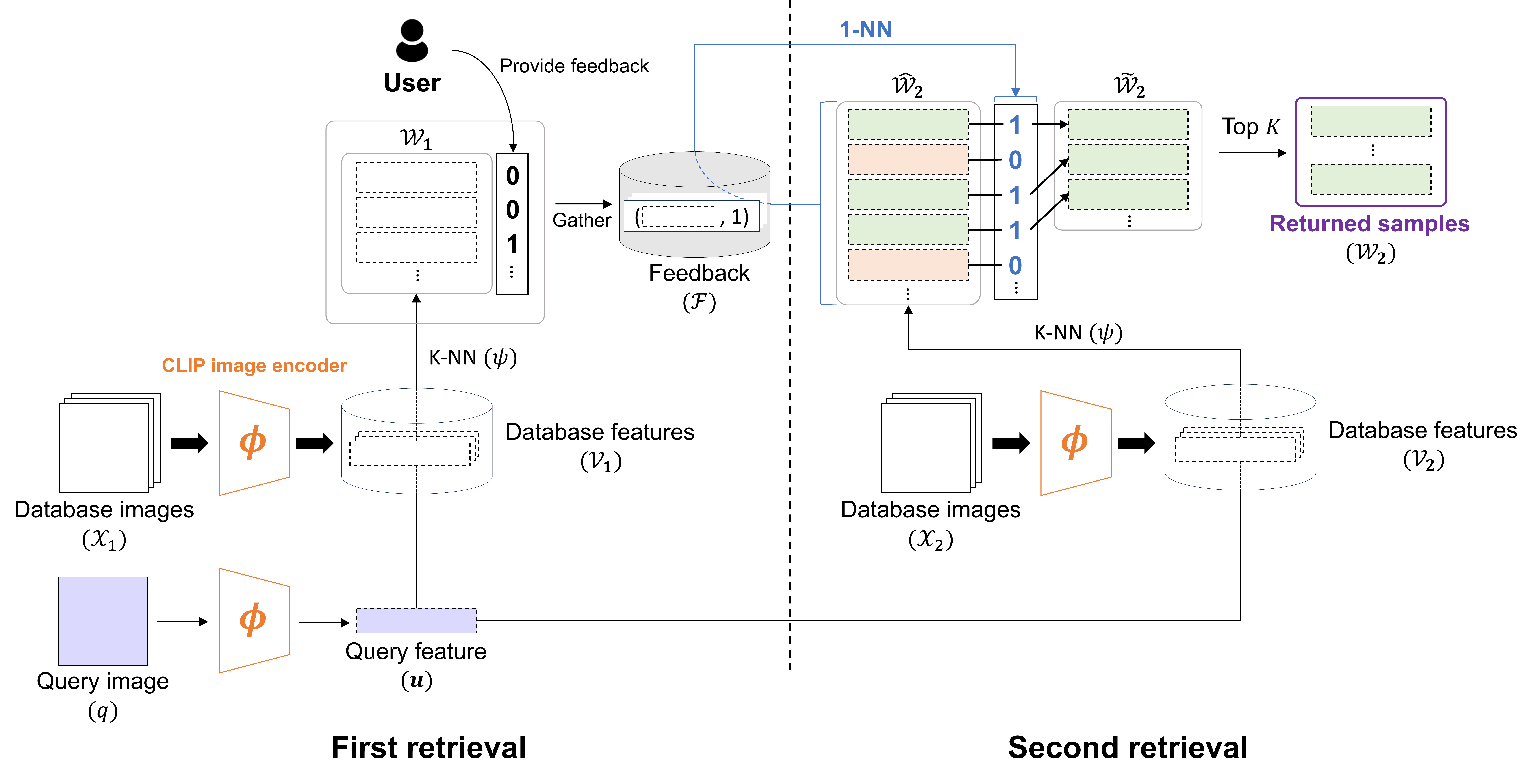}
    \caption{Overall of our proposed method. The updated retrieval $\tilde{\psi}$ is a retrieval algorithm that retrieves and returns $\mathcal{W}_2$ from $\mathcal{V}_2$. }
    \label{fig:proposed_method_all}
\end{figure}

\cref{fig:proposed_method_all} illustrates the overview of our proposed method. We propose a CLIP-based retrieval system with relevance feedback. When updating the retrieval algorithm, our retrieval system predicts whether each sample $\mathbf{w} \in \mathcal{W}_2$ is preferable for the user according to the information in $\mathcal{F}$.

We use CLIP image encoder as our retrieval system's encoder $\mathbf{\phi}$. We aim to update the retrieval algorithm for various datasets and users' preferences, so we must extract appropriate information from images with an image encoder. 
Therefore, we utilize off-the-shelf CLIP because CLIP has a rich semantic space and contains knowledge from various types of data~\cite{sb-zsir}. Several previous works use CLIP to handle zero-shot image recognition tasks. For example, Nakata \etal~\cite{revisiting-knn} uses a CLIP image encoder to achieve zero-shot classification, and Vinker \etal~\cite{clipasso} proposes a zero-shot object sketching method utilizing both CLIP image encoder and CLIP text encoder. We follow this line and adopt an off-the-shelf CLIP image encoder as a feature extractor to construct an image retrieval system without additional training.

After the first retrieval with the feedback $\mathcal{F}$, our retrieval system prepares a binary classifier $f: \mathbb{R}^D \to \{0, 1\}$ to predict the user's preference. There could be several options for $f$, but in our case, $f$ is a simple 1-NN classifier over $\mathcal{F}$: 
\begin{equation}
    f(\mathbf{a}) = b_* ~~ \mathrm{where} ~~ (\mathbf{y}_*, b_*) = \psi(\mathbf{a}, 1, \mathcal{F}),
\end{equation}
Here, we use a slight abuse notation for $\psi$. This form of $f$ enables us to update the retrieval algorithm online without requiring any additional training time.

In the second retrieval phase, our retrieval system utilizes $f$ to execute the updated retrieval $\tilde{\psi}$ described in \cref{eq:updated-retrieval-definition}. First, our retrieval system executes K-NN retrieval to obtain the candidates of the final results as follows:
\begin{equation}\label{eq:naive-knn-second}
    \hat{\mathcal{W}}_2 = \psi(\mathbf{u}, \hat{K}, \mathcal{V}_\mathrm{2}).
\end{equation}
Here, $\hat{K}$ is a sufficiently large value. Our retrieval system then refines $\hat{\mathcal{W}}_2$ by asking $f$ to obtained the refined candidates $\tilde{\mathcal{W}}_2$ as \cref{eq:f}:
\begin{equation}\label{eq:f}
    \tilde{\mathcal{W}}_2 = \{ \mathbf{w}_2 \in \hat{\mathcal{W}}_2
    \mid f(\mathbf{w}_2) = 1 \}.
\end{equation}
Finally, our retrieval system picks up the top $K$ elements.
\begin{equation}\label{eq:w1-k}
    \mathcal{W}_2 =  \tilde{\mathcal{W}}_2[1:K].
\end{equation}
We put \cref{eq:naive-knn-second,eq:f,eq:w1-k} all together to implement $\tilde{\psi}$ in \cref{eq:updated-retrieval-definition}.

We can consider the above procedures to be the simplest form of relevance feedback. Our work finds that the CLIP image encoder achieves surprising retrieval accuracy with such simple relevance feedback.
% \matsui{
% \begin{equation}
%         \mathcal{W}_2 = \tilde{\psi}(\mathbf{u}, K, \mathcal{V}_2) \coloneq  \{ \mathbf{w}_2 \in \psi(\mathbf{u}, \hat{K}, \mathcal{V}_\mathrm{2}) \mid f(\mathbf{w}_2) = 1 \}[1:K].
% \end{equation}
% }

\begin{figure}[tb]
    \centering
    \includegraphics[width=0.9\linewidth]{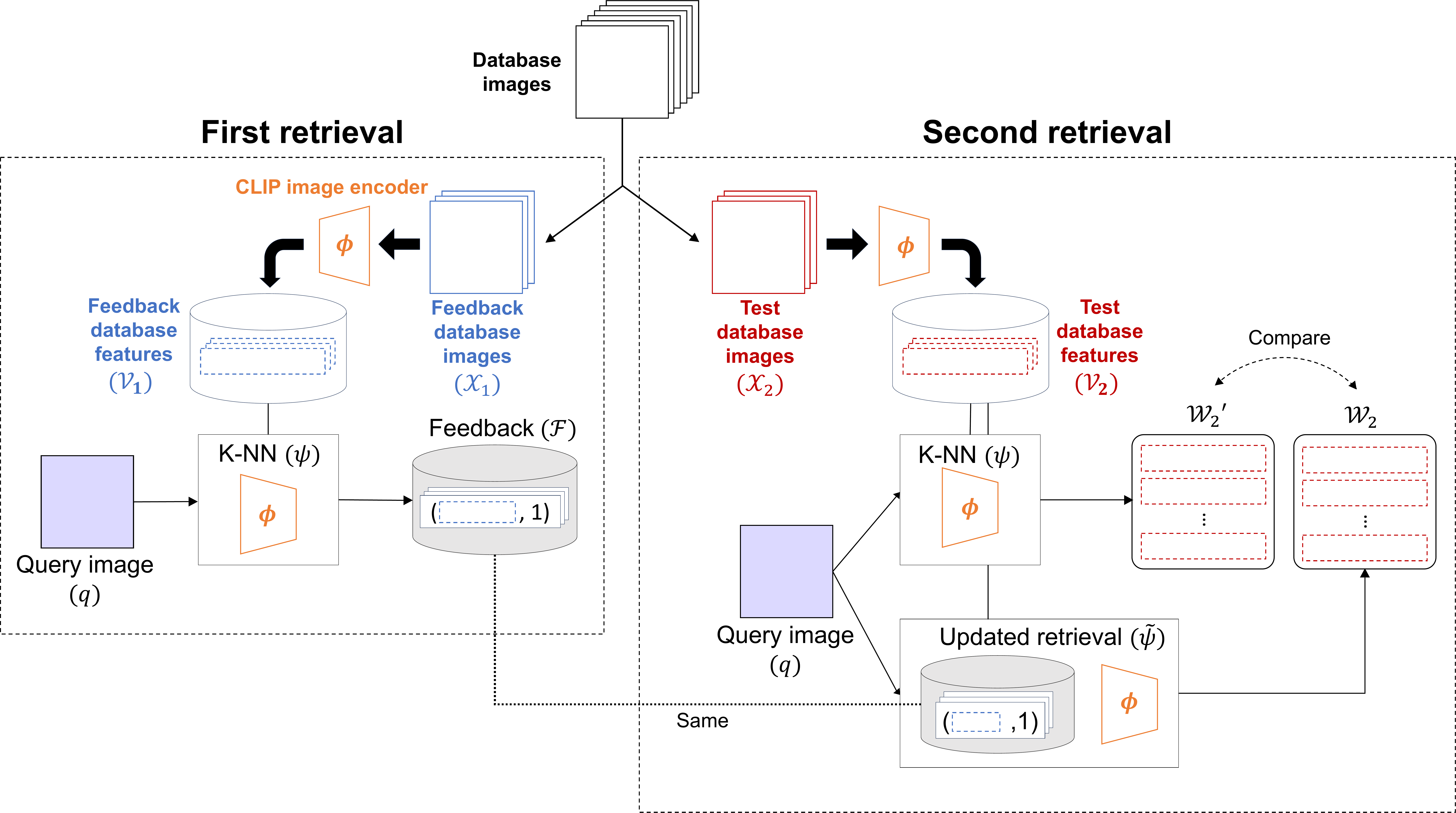}
    \caption{Evaluation of retrieval algorithm with relevance feedback. We omit the detail of the second retrieval process described in \cref{fig:proposed_method_all}. }
    \label{fig:test_and_control_method}
\end{figure}

\subsection{Evaluation}\label{subsec:approach-evaluation}

\cref{fig:test_and_control_method} illustrates the overall evaluation framework.
As we explained, in the practical application of relevance feedback, the database features in the first and second retrieval are the same: $\mathcal{X}_1 = \mathcal{X}_2$ and $\mathcal{V}_1 = \mathcal{V}_2$. However, we could not correctly compare the performance of $\psi$ and $\tilde{\psi}$ in this way, because our retrieval system knows the ground-truth label of whether the user prefers each of the top-$M$ returned samples in the database of the first retrieval.

To evaluate the update of retrieval algorithm performance correctly, we revisit the classic evaluation method of relevance feedback, called the test and control method~\cite{relevance-feedback-evaluation-survey,chang1971evaluation}. First, we split all the database images into two subsets: \textbf{feedback-database-images} and \textbf{test-database-images}. 
In the first retrieval, we use the feedback-database-images to obtain samples used for relevance feedback. In the second retrieval, we use the test-database-images to evaluate retrieval accuracy. That is, the feedback-database-images are $\mathcal{X}_1$, and the test-database-images are $\mathcal{X}_2$. Here, $\mathcal{X}_1 \cap \mathcal{X}_2 = \emptyset$.

After splitting, we construct $\mathcal{V}_1$ and $\mathcal{V}_2$ as described in \cref{subsec:approach-problem-settings}. To compare the retrieval performance of $\psi$ and $\tilde{\psi}$, we execute retrieval for $\mathcal{V}_2$ with $\psi$.
\begin{equation}\label{eq:w_control}
    \mathcal{W}_2' = \psi(\mathbf{u}, K, \mathcal{V}_2).
\end{equation}
$\mathcal{W}_2'$ means samples returned from $\mathcal{V}_2$ with the simple K-NN. Finally, we compare $\mathcal{W}_2$ and $\mathcal{W}_2'$ and measure whether the returned samples are refined.

When we execute the splitting, we set the size of the feedback and test database to the same. Also, we adopt a stratified sampling approach, where the ratios of the labels in each subset are equal. We split the dataset in this way because the test and control method is unbiased only if the two database datasets have equivalent numbers and distributions of samples, as Hull~\cite{relevance-feedback-evaluation-survey} says.

\section{Experiment}\label{sec:experiment}

Although our retrieval system requires users’ feedback, we automatically generate pseudo-feedback for evaluation purposes. We use labeled datasets to automatically generate a user’s feedback for each returned sample. In this work, we assume that a user could provide binary feedback correctly. That is, we generate positive feedback for positive samples and negative feedback for negative ones.

In our experiment, we consider evaluating accuracy of our retrieval algorithm with relevance feedback. First, we consider category-based image retrieval (\cref{subsec:category-based}), where a user prefers samples with the same label as the query image. Next, we consider two retrieval settings which considers diverse user preferences: one-label-based image retrieval (\cref{subsec:one-label-based}) and conditioned image retrieval (\cref{subsec:conditioned}). We devise the two retrieval tasks from the same motivation as the GeneCIS benchmark~\cite{genecis}, where users have various intentions.

Throughout all our experiments, we split the evaluation dataset into three groups: query images, feedback-database-images $\mathcal{X}_1$, and test-database-images $\mathcal{X}_2$. In all experimental settings, the size ratio of these groups is $1:2:2$. We conduct retrieval tasks on ten different splittings and calculate the average and standard deviation of $\mathrm{Recall@K}$. We execute all experiments in a single Tesla V100 GPU. Images are resized to $256 \times 256$ and then cropped to $224 \times 224$ at the center to input them into the model. We set $\hat{K} = |\mathcal{V}_2|$\footnote{Here, setting $\hat{K} = |\mathcal{V}_2|$ means that our retrieval system checks all samples in the test-database-images. This setting might seem to increase the retrieval runtime too much, but our experiments reveal that this is not the case. We discuss this point in the supplementary material.}. Furthermore, if we could not obtain any returned samples' candidates $\tilde{\mathcal{W}}$ for one query, we view this trial as a failure and calculate $\mathrm{Recall@K}$ as $0$.

In all experiments in \cref{sec:experiment}, we fix the feedback size $M$ to $50$. From our conversations with industrial engineers, $M=50$ is found to be within optimal values for many industrial applications such as defects analysis. It is because $M=50$ is large enough to provide useful macro analysis in a single round as engineers need to check the retrieval results anyway, and small enough for feedbacks creation. In addition, as an example, by defaulting feedback labels to ``False'', only $8.62$ operations in average are needed to flip positive samples to ``True'' in our experiment (CUB-200-2011) conducted in \cref{subsec:category-based}. Furthermore, the binary nature of the feedback allows more efficient feedback methods, such as signals from user’s brain activity.

\subsection{Category-based Image Retrieval}\label{subsec:category-based}

Category-based image retrieval is a common retrieval task. We calculate $\mathrm{Recall@K}$ based on whether each returned sample's label is identical to the query's. After the first retrieval, we generate binary feedback for each returned sample based on whether its label is identical to the query's. We use two datasets: CUB-200-2011~\cite{CUB200} and Cars-196~\cite{cars196}. Each sample in the two datasets has a single label.

We compare $\mathrm{Recall@K}$ of our retrieval system with metric-learning-based image retrieval. We choose triplet loss~\cite{triplet} as simple metric learning and HIST~\cite{hist} as state-of-the-art metric learning. 
As a metric-learning-specific procedure, we follow the common practice of metric learning~\cite{metric-reality-check} and split the whole labels into two groups in advance: the training labels and the evaluation labels. Image encoders for metric learning methods are trained with images of the training labels. That is all for the metric-learning-specific procedure, and we handle only the remaining images (with the evaluation labels) as an evaluation dataset below. We execute the dataset splitting and obtain query images, feedback-database-images $\mathcal{X}_1$, and test-database-images $\mathcal{X}_2$. We use the same database images and query images to evaluate each retrieval system.
When calculating $\mathrm{Recall@K}$ of metric-learning-based image retrieval, we encode test-database-images $\mathcal{X}_2$ with the trained encoder and construct test database features. Next, we encode each query image $q$ into a $D$-dimensional feature with the same encoder, execute K-NN retrieval, and calculate $\mathrm{Recall@K}$.

We use ResNet-50 as a backbone of all image encoders to compare $\mathrm{Recall@K}$ equally. To obtain a visual encoder trained with triplet loss~\cite{triplet}, we execute training ourselves following the implementation of prior work~\cite{proxy-anchor}. For HIST~\cite{hist}, the authors publicize their pre-trained models to each dataset, so we use them in our experiment.

\subsubsection*{Result}

\begin{table}[tb]
    \centering
    \caption{$\mathrm{Recall@K}$ in category-based image retrieval. We use ResNet-50 as an encoder architecture. Note that $\mathrm{Recall@K}$ of HIST~\cite{hist} differs from that reported in the original paper because the experimental condition differs. }
        \begin{tabular}{@{}llllllll@{}} \toprule
            Dataset & Method  & Feedback & Training  & $K=1$     & $K=2$     & $K=4$     & $K=8$    \\ \midrule
            CUB-200-2011& CLIP   &          &          & $46.2$\fontsize{7}{7}$\pm1.1$&$59.9$\fontsize{7}{7}$\pm0.8$&$73.3$\fontsize{7}{7}$\pm1.1$& $84.4$\fontsize{7}{7}$\pm0.9$  \\
            & Triplet  &  &  \checkmark &$61.7$\fontsize{7}{7}$\pm0.6$&$73.0$\fontsize{7}{7}$\pm0.3$&$82.7$\fontsize{7}{7}$\pm0.4$&$90.1$\fontsize{7}{7}$\pm0.3$\\
            & HIST        &   & \checkmark  &$\bm{67.9}$\fontsize{7}{7}$\pm0.9$&$78.4$\fontsize{7}{7}$\pm0.8$& $86.4$\fontsize{7}{7}$\pm0.8$ & $91.9$\fontsize{7}{7}$\pm0.4$ \\
             &  Ours & \checkmark   &          & $64.2$\fontsize{7}{7}$\pm1.1$& $\bm{78.6}$\fontsize{7}{7}$\pm1.1$&$\bm{88.5}$\fontsize{7}{7}$\pm0.7$& $\bm{93.8}$\fontsize{7}{7}$\pm0.4$  \\
            \midrule
            Cars-196&CLIP    &          &        & $69.5$\fontsize{7}{7}$\pm1.0$   & $80.8$\fontsize{7}{7}$\pm1.0$ & $89.1$\fontsize{7}{7}$\pm0.8$&$94.6$\fontsize{7}{7}$\pm0.6$\\
            &Triplet   &          &  \checkmark  & $83.4$\fontsize{7}{7}$\pm0.8$&$90.0$\fontsize{7}{7}$\pm0.9$&$94.2$\fontsize{7}{7}$\pm1.0$&$96.7$\fontsize{7}{7}$\pm1.1$\\
            & HIST  &          & \checkmark&$85.2$\fontsize{7}{7}$\pm0.9$&$91.0$\fontsize{7}{7}$\pm0.6$&$94.6$\fontsize{7}{7}$\pm0.4$& $97.0$\fontsize{7}{7}$\pm0.2$  \\
             & Ours &    \checkmark    &        &$\bm{86.4}$\fontsize{7}{7}$\pm0.6$&$\bm{94.7}$\fontsize{7}{7}$\pm0.5$&$\bm{98.0}$\fontsize{7}{7}$\pm0.5$&$\bm{99.0}$\fontsize{7}{7}$\pm0.3$  \\
            \bottomrule
        \end{tabular}
    \label{tbl:cbir-comparision}
\end{table}

\cref{tbl:cbir-comparision} shows the results. Our retrieval system utilizes relevance feedback and achieves competitive results with state-of-the-art metric learning methods, despite not training image encoders specifically for each dataset. $\mathrm{Recall@1}$ of HIST is higher than that of our retrieval system in both datasets, but our retrieval system surpasses HIST in $\mathrm{Recall@2,4,8}$ in both datasets. 

These results mean that simple user feedback for each query is enough to enable the CLIP image encoder to achieve competitive retrieval accuracy with metric learning methods. This experiment reveals that we can build accurate retrieval systems by combining CLIP and realistic feedback size.

\subsection{One-label-based Image Retrieval}\label{subsec:one-label-based}

One-label-based image retrieval is an experimental setting simulating a user focusing on one of the query image's characteristics. Each user expects that the returned samples have the same characteristics on which the user focuses. For example, consider a user providing a query image of a boy with a hat standing before a door. The user hopes to obtain a hat image, not paying attention to other characteristics such as a door and a boy. In this case, the user provides positive feedback to images of a hat, and our retrieval aims to adapt to the user's preference and return images of a hat. Metric learning cannot handle this task because it does not take each user's preference as input.

To simulate this setting, we use three datasets in which each image has multiple labels: MIT-States~\cite{MITStates}, Fashion200k~\cite{fashion200k}, and COCO 2017 Panoptic Segmentation~\cite{coco-panoptic, coco}. For each query image, we focus on one of the query image's labels. We view the label as the characteristic each user focuses on, and regard samples with the focused label as positive when generating user feedback and calculating $\mathrm{Recall@K}$. For example, when a query image has labels of ``boy,'' ``hat,'' and ``door,'' and we focus on the label ``hat,'' we simulate a user who provides the query image aiming to obtain images of a hat. In this case, we generate positive feedback to samples with the label ``hat'' and regard such samples as positive when calculating $\mathrm{Recall@K}$.

For each query image $q$ that has $L$ labels $\{l_i\}_{i=1}^{L}$, we execute trials $L$ times. In $i$\textsuperscript{th} retrieval, we focus on the label $l_i$, generate user feedback, and calculate $\mathrm{Recall@K}$ as explained above.

% \subsubsection*{Implementation Detail}
We use test images in MIT-States~\cite{MITStates} and Fashion200k~\cite{fashion200k} and evaluation data of COCO 2017 Panoptic Segmentation~\cite{coco-panoptic, coco}. We view each dataset as an evaluation dataset and execute the dataset splitting. We regard attributes and nouns for each image of MIT-States and Fashion200k as labels, and annotated objects of each COCO 2017 Panoptic Segmentation image as labels. As backbones of CLIP image encoders, we choose ViT-B/32 and ResNet-50.

\subsubsection*{Result}

\begin{table}[tb]
    \caption{$\mathrm{Recall@K}$ of one-label-based image retrieval with CLIP image encoder.}
    \centering
    \begin{tabular}{@{}lllllll@{}} \toprule
    Dataset & Arch  & Feedback & $K=1$   & $K=2$     & $K=4$     & $K=8$    \\ \midrule
     Fashion200k & ViT-B/32 &    & $67.9$\fontsize{7}{7}$\pm0.8$   & $75.8$\fontsize{7}{7}$\pm0.8$&$82.4$\fontsize{7}{7}$\pm0.8$& $87.5$\fontsize{7}{7}$\pm0.8$ \\
          &    &\checkmark & $\bm{76.1}$\fontsize{7}{7}${\pm0.6}$&$\bm{83.5}$\fontsize{7}{7}${\pm0.5}$&$\bm{87.8}$\fontsize{7}{7}${\pm0.4}$& $\bm{90.1}$\fontsize{7}{7}${\pm0.5}$ \\ 
    \cmidrule(){2-7}
      & R50 &   & $66.1$\fontsize{7}{7}$\pm0.6$   & $74.5$\fontsize{7}{7}$\pm0.8$&$81.2$\fontsize{7}{7}$\pm0.6$& $86.6$\fontsize{7}{7}$\pm0.6$ \\
      &   &\checkmark   & $\bm{74.3}$\fontsize{7}{7}${\pm0.9}$&$\bm{82.3}$\fontsize{7}{7}${\pm0.7}$&$\bm{87.2}$\fontsize{7}{7}${\pm0.5}$& $\bm{89.6}$\fontsize{7}{7}${\pm0.4}$ \\ \midrule
      MIT States & ViT-B/32 &   & $40.4$\fontsize{7}{7}$\pm0.4$&$51.8$\fontsize{7}{7}$\pm0.2$&$62.7$\fontsize{7}{7}$\pm0.2$& $73.1$\fontsize{7}{7}$\pm0.2$ \\
       &    & \checkmark  & $\bm{49.9}$\fontsize{7}{7}${\pm0.6}$&$\bm{63.5}$\fontsize{7}{7}${\pm0.7}$&$\bm{74.3}$\fontsize{7}{7}${\pm0.5}$& $\bm{81.6}$\fontsize{7}{7}${\pm0.4}$ \\ 
         \cmidrule(){2-7}
        & R50 &     & $37.9$\fontsize{7}{7}$\pm0.2$ & $49.0$\fontsize{7}{7}$\pm0.2$&$60.2$\fontsize{7}{7}$\pm0.3$& $70.7$\fontsize{7}{7}$\pm0.3$ \\
         &   &\checkmark  & $\bm{47.1}$\fontsize{7}{7}${\pm0.4}$&$\bm{60.4}$\fontsize{7}{7}${\pm0.4}$&$\bm{71.7}$\fontsize{7}{7}${\pm0.3}$& $\bm{79.4}$\fontsize{7}{7}${\pm0.3}$ \\\midrule
       COCO & ViT-B/32 &  & $49.3$\fontsize{7}{7}$\pm0.8$&$63.1$\fontsize{7}{7}$\pm0.7$&$75.6$\fontsize{7}{7}$\pm0.6$& $85.0$\fontsize{7}{7}$\pm0.4$ \\
        &  &\checkmark  & $\bm{58.3}$\fontsize{7}{7}${\pm0.8}$&$\bm{73.7}$\fontsize{7}{7}${\pm0.9}$&$\bm{85.6}$\fontsize{7}{7}${\pm0.6}$& $\bm{92.7}$\fontsize{7}{7}${\pm0.4}$ \\ 
         \cmidrule(){2-7}
        &R50 &  & $49.7$\fontsize{7}{7}$\pm0.8$ & $63.3$\fontsize{7}{7}$\pm0.5$&$75.4$\fontsize{7}{7}$\pm0.6$& $84.8$\fontsize{7}{7}$\pm0.4$ \\
         &   &\checkmark & $\bm{58.4}$\fontsize{7}{7}${\pm1.0}$&$\bm{73.8}$\fontsize{7}{7}${\pm0.7}$&$\bm{85.5}$\fontsize{7}{7}${\pm0.6}$& $\bm{92.5}$\fontsize{7}{7}${\pm0.5}$ \\
      \bottomrule
    \end{tabular}
    \label{tbl:one-label-based}
\end{table}

\begin{figure}[tb]
    \centering
    \includegraphics[width=0.85\linewidth]{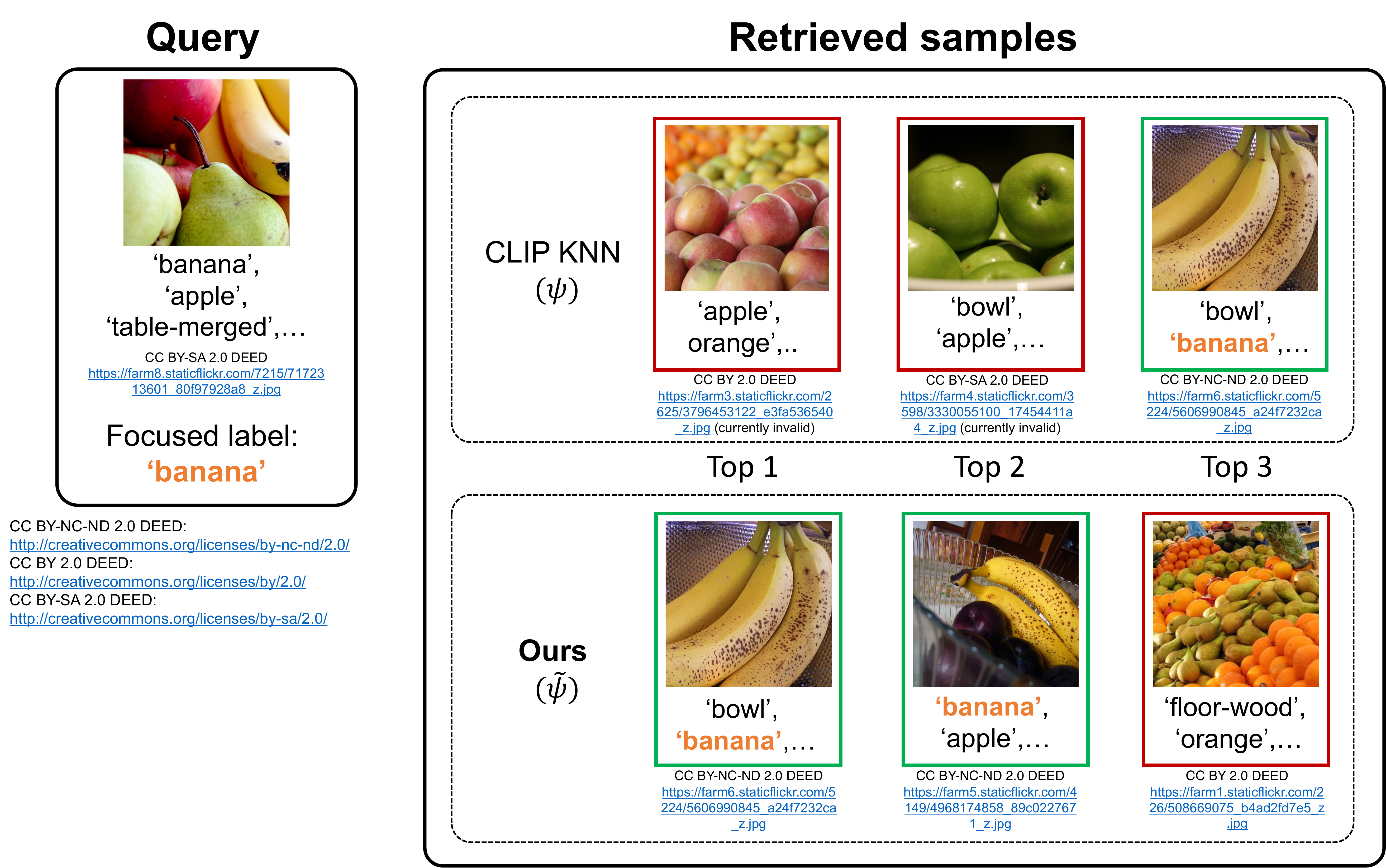}
    \caption{An example of one-label-based image retrieval in COCO.}
    \label{fig:one-label-coco}
\end{figure}

\cref{tbl:one-label-based} shows the results. Our retrieval system successfully improves $\mathrm{Recall@K}$ with relevance feedback in all datasets and encoder architectures by a significant margin. With the relevance feedback, $\mathrm{Recall@1}$ improves by up to $9.5\%$. 
These results mean that our retrieval system successfully adapts to each user's preference in our simulation experiment. \cref{fig:one-label-coco} illustrates one example of our retrieval system's trial with CLIP ViT-B/32 in COCO dataset. In this case, we focus on a label ``banana,'' simulating a user who provides the query image in \cref{fig:one-label-coco} searching for banana images. With relevance feedback, our retrieval system can return images with the label ``banana'' more accurately.

Regarding the encoder architecture, ViT-B/32 surpasses ResNet-50 in $\mathrm{Recall@K}$. In terms of $\mathrm{Recall@1}$, ViT-B/32 ourperforms ResNet-50 by $1.8\%$ in Fashion200k, and $2.1\%$ in MIT-States. These results suggest that CLIP image encoders with large architectures work better in one-label-based image retrieval settings.

\subsection{Conditioned Image Retrieval}\label{subsec:conditioned}

Conditioned image retrieval is an experimental setting simulating a user who searches for similar images to a query image but differs from it in some aspects. For example, a user provides a ripe apple image, but they want unripe apple images. In this case, the user provides positive feedback to images of an unripe apple, and our retrieval system aims to adapt to the user's preference and return unripe apple images. Metric learning cannot handle this task because it does not take such users' preferences as input.

To simulate this setting, we follow previous conditioned image retrieval studies~\cite{compose-ae,tirg, genecis} and use MIT-States~\cite{MITStates}. Each image in MIT-States has one adjective label and one noun label. 
We use the same experimental settings as previous conditioned image retrieval studies. For each query image $q$ with the adjective label $l_{\mathrm{adj}}$ and noun label $l_{\mathrm{noun}}$, we choose one adjective label $l_{\mathrm{adj}}'$ that is different from $l_{\mathrm{adj}}$. We view images that have both $l_{\mathrm{adj}}'$ and $l_{\mathrm{noun}}$ as those the user prefers, and regard samples with $l_{\mathrm{adj}}'$ and $l_{\mathrm{noun}}$ as positive when generating and calculating $\mathrm{Recall@K}$. For example, when a query image has $l_{\mathrm{adj}}=~$``ripe'' and $l_{\mathrm{noun}}=~$``apple,'' and we choose $l_{\mathrm{adj}}'=~$``unripe,'' we simulate a user who provides a ripe apple image aiming to obtain images of an unripe apple. In this case, we automatically generate positive feedback to samples with labels ``unripe'' and ``apple,'' and regard such samples as positive when calculating $\mathrm{Recall@K}$.

For each $q$ with $l_{\mathrm{adj}}$ and $l_{\mathrm{noun}}$, we try all possible adjective labels $\mathcal{L}_{\mathrm{adj}} \backslash \{ l_{\mathrm{adj}} \}$. Here, $\mathcal{L}_{\mathrm{adj}}$ is a set of all adjective labels. When we define $\{ l_{\mathrm{adj}}^{i} \}_{i=1}^{N} \coloneq \mathcal{L}_{\mathrm{adj}} \backslash \{ l_{\mathrm{adj}} \}~(N=|\mathcal{L}_{\mathrm{adj}} \backslash \{ l_{\mathrm{adj}} \}|)$, we execute trials $N$ times for $q$. In $i$\textsuperscript{th} retrieval, we choose $l_{\mathrm{adj}}^{i}$, generate binary feedback, and calculate $\mathrm{Recall@K}$ as explained above.

% \subsubsection*{Comparison with Multimodal Retrieval}
Additionally, we compare our retrieval system with multimodal retrieval. We choose GeneCIS~\cite{genecis} as the state-of-the-art multimodal retrieval addressing conditioned image retrieval on MIT-States. GeneCIS consists of visual encoder ${\mathbf{\phi}'_v}:\mathcal{I} \to \mathbb{R}^D$ and multimodal encoder $\mathbf{\phi}'_{\mathrm{mm}}: \mathcal{I} \times \mathcal{T} \to \mathbb{R}^D$. Here, $\mathcal{T}$ is a set of all text. 
We reimplement GeneCIS and calculate $\mathrm{Recall@K}$ in the same query images and test-database-images $\mathcal{X}_2$ as those used to evaluate our retrieval system. We obtain test database features as $\mathcal{V}_2' \coloneq \{\mathbf{\phi}'_v(x) \mid x \in \mathcal{X}_2\}$.
Consider the query image $q$ has $l_{\mathrm{adj}}$ and $l_{\mathrm{noun}}$, and we choose $l_{\mathrm{adj}}'(\neq l_{\mathrm{adj}})$. We generate the query feature $\mathbf{u}_{\mathrm{mm}} \coloneq \mathbf{\phi}'_{\mathrm{mm}}(q, l_{\mathrm{adj}}')$. Note that each adjective label is represented as text: $l_{\mathrm{adj}}' \in \mathcal{T}$. We execute retrieval and return samples $\mathcal{W}_{\mathrm{mm}}$ as follows:
\begin{equation}
    \mathcal{W}_{\mathrm{mm}} = \psi(\mathbf{u}_{\mathrm{mm}}, K, \mathcal{V}_2').
\end{equation}

% \subsubsection*{Implementation Detail}
We view test images of MIT-States as an evaluation dataset and execute the dataset splitting. We choose ResNet-50x4 and ViT-B/16 as CLIP image encoder backbones. The authors publicize the pre-trained models of GeneCIS~\cite{genecis}, and we use them for comparison.

\subsubsection*{Result}

\begin{table}[tb]
    \centering
    \caption{$\mathrm{Recall@K}$ of conditioned image retrieval for MIT States. Note that $\mathrm{Recall@K}$ of GeneCIS~\cite{genecis} differs from that reported in the original paper because the experimental condition differs.}
    \begin{tabular}{@{}lllllll@{}} \toprule
        Arch & Method    & Feedback  & $K=1$   & $K=2$     & $K=4$     & $K=8$    \\ \midrule
        ViT-B/16 & CLIP & -  &  $4.28$\fontsize{7}{7}$\pm0.14$&$7.99$\fontsize{7}{7}$\pm0.23$&$14.1$\fontsize{7}{7}$\pm0.3$& $23.4$\fontsize{7}{7}$\pm0.3$ \\ 
        & GeneCIS & Text & $\bm{17.3}$\fontsize{7}{7}${\pm0.6}$  & $\bm{26.7}$\fontsize{7}{7}${\pm0.7}$ & $\bm{38.0}$\fontsize{7}{7}${\pm0.7}$&$\bm{50.7}$\fontsize{7}{7}${\pm0.7}$ \\
         & Ours  & Binary  & $\bm{17.3}$\fontsize{7}{7}${\pm0.9}$&$\bm{26.7}$\fontsize{7}{7}${}$\fontsize{7}{7}${\pm1.1}$&$36.1$\fontsize{7}{7}$\pm1.0$& $42.7$\fontsize{7}{7}$\pm0.9$ \\
        \midrule
        R50x4 & CLIP &  -  & $4.16$\fontsize{7}{7}$\pm0.08$ & $7.72$\fontsize{7}{7}$\pm1.2$&$13.7$\fontsize{7}{7}$\pm0.2$& $22.9$\fontsize{7}{7}$\pm0.2$ \\
        & GeneCIS   &   Text   & $15.7$\fontsize{7}{7}$\pm1.0$&$24.2$\fontsize{7}{7}$\pm1.9$&$34.8$\fontsize{7}{7}$\pm1.0$& $\bm{47.2}$\fontsize{7}{7}${\pm0.8}$ \\
         & Ours & Binary & $\bm{16.4}$\fontsize{7}{7}${\pm0.8}$&$\bm{25.8}$\fontsize{7}{7}${\pm1.0}$&$\bm{35.0}$\fontsize{7}{7}${\pm0.8}$& $41.6$\fontsize{7}{7}$\pm0.8$ \\ 
      \bottomrule
    \end{tabular}
    \label{tab:conditioned-image-retrieval}
\end{table}

\cref{tab:conditioned-image-retrieval} shows the results. We successfully improve $\mathrm{Recall@K}$ with relevance feedback from simple CLIP K-NN retrieval without relevance feedback by a large margin. Furthermore, when we provide a realistic amount of relevant feedback ($M=50$), our retrieval system achieves competitive accuracy with state-of-the-art multimodal retrieval. $\mathrm{Recall@1}$ of ours and GeneCIS are almost equal in both encoder architectures. These results mean that our retrieval system adapts to each user's preference without textual information in this setting.

\section{Additional Analysis}

\subsection{Architectures of CLIP Image Encoders and the Feedback Size}\label{subsec:various-encoder-and-feedback-size}

\begin{figure}[tb]
    \centering
    \includegraphics[width=0.85\linewidth]{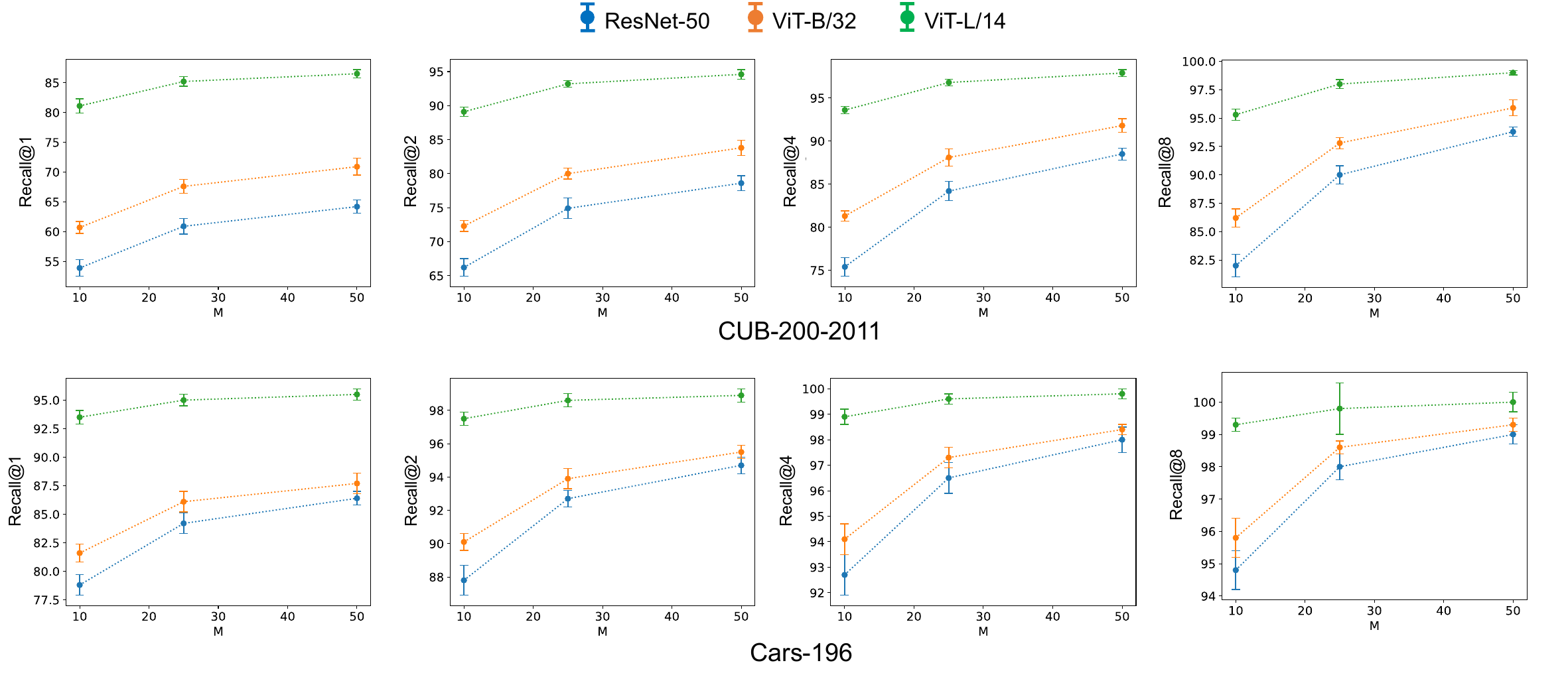}
    \caption{Comparison among various kinds of CLIP and $M$.}
    \label{fig:various-clip-cbir}
\end{figure}

In category-based image retrieval settings, we change the architecture of CLIP image encoder and evaluate how our retrieval system improves the retrieval accuracy. We choose ResNet-50, ViT-B/32, and ViT-L/14 as backbone architectures and calculate $\mathrm{Recall@K}$ in each setting. Also, we vary the feedback size ($M$) and assess the relationship between the feedback size and the retrieval accuracy. We vary $M$ among $10$, $25$, and $50$.

\cref{fig:various-clip-cbir} shows the results of comparison among various kinds of CLIP and $M$. Larger $M$ leads to better $\mathrm{Recall@K}$. We can infer that our retrieval system obtains more information from the larger feedback. Moreover, the larger architecture of CLIP image encoder achieves higher accuracy, implying that the larger CLIP image encoder extracts more appropriate information from images. 

In particular, CLIP ViT-L/14 performs well all the way down towards $M=10$. Note that as an example for CUB-200-2011 experiment with this model, by defaulting feedback labels to ``False'', only $13.8,10.4,5.7$ operations in average are needed to flip positive samples to ``True'' for $M=50,25,10$ respectively.

\subsection{Retrieval Accuracy and the Number of Positive Feedback}

We examine the influence of positive samples in the first retrieval on the retrieval performance in the second retrieval. We choose category-based image retrieval settings, where the architecture of CLIP image encoder is ViT-B/32 and $M=50$. We count the positive feedback in the first retrieval, which is represented as $| \{  (\mathbf{w}, b) \in \mathcal{F} \mid b = 1 \} |$. At the same time, we calculate $\mathrm{MAP@R}$ of the returned samples in the second retrieval. $\mathrm{MAP@R}$ is calculated as follows~\cite{metric-reality-check}:
\begin{align}
     \mathrm{MAP@R} &= \frac{1}{R}\sum_{i=1}^{R}P(i).\\
        P(i) &= \begin{cases}
    \mathrm{precision~at~}i~(\text{if the $i$\textsuperscript{th} returned sample is positive}),\\
    0~(\text{otherwise}).
    \end{cases}
\end{align}
Here, $R$ is the total number of positive samples in $\mathcal{X}_2$. We execute the retrieval for all query images and collect pairs of the number of positive feedback in the first retrieval and $\mathrm{MAP@R}$ of the second retrieval.

\begin{figure}[tb]
    \centering
    \begin{subfigure}{0.4\linewidth}
        \centering
        \includegraphics[width=0.8\linewidth]{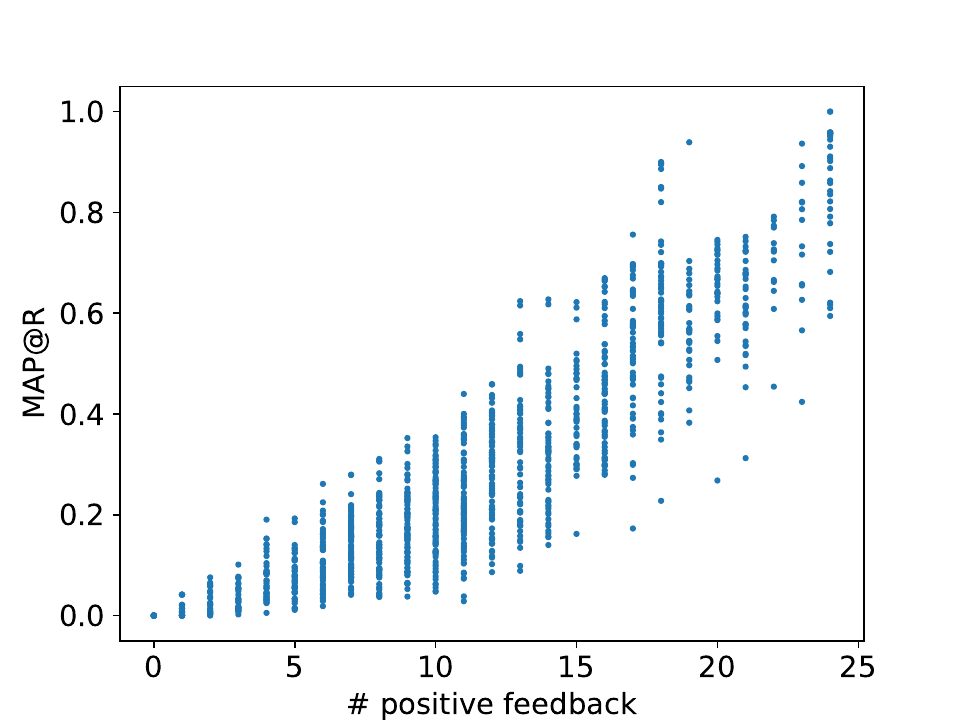}
        \caption{CUB-200-2011}
    \end{subfigure}
    \hfill
    \begin{subfigure}{0.4\linewidth}
    \centering
        \includegraphics[width=0.8\linewidth]{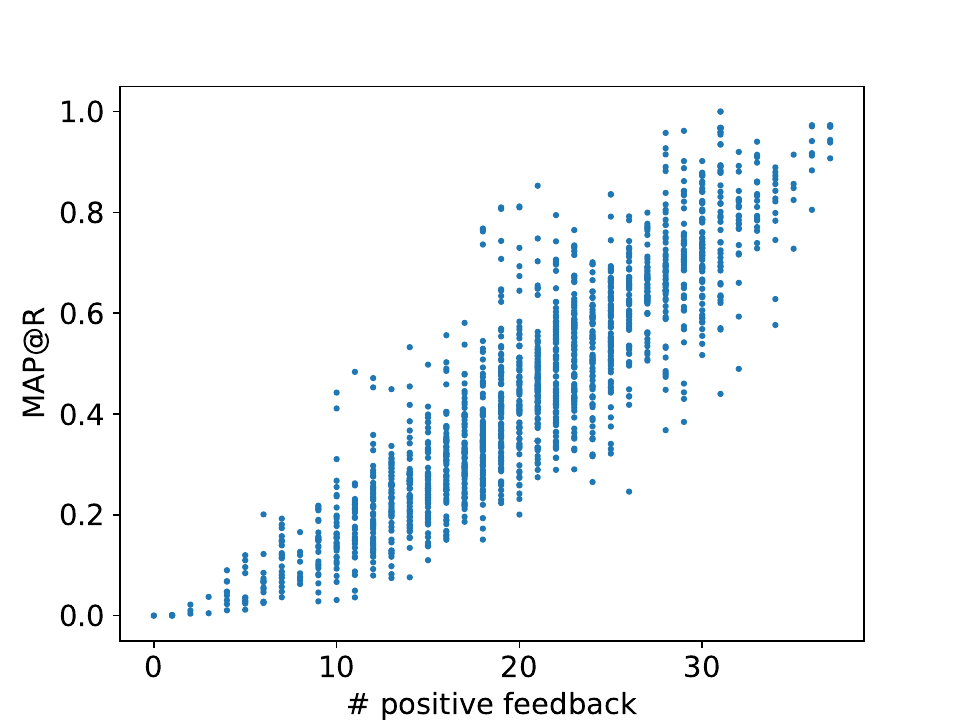}
        \caption{Cars-196}
    \end{subfigure}
    \caption{Relashinship between the number of positive feedback and $\mathrm{MAP@R}$.}
    \label{fig:positive-feedback-and-ap}
\end{figure}

\cref{fig:positive-feedback-and-ap} shows the results for each dataset. Each data point represents one query. A positive correlation exists between the number of positive feedback and $\mathrm{MAP@R}$. These results suggest that more positive feedback leads to higher retrieval performance. When we have positive samples in the user feedback, we can accurately predict whether each sample in the second retrieval is preferable.

\subsection{Retrieval Runtime}\label{subsec:retrieval-time}

We examine the retrieval runtime of our retrieval system. We choose category-based image retrieval settings. We choose ViT-B/32 as the backbone architecture of CLIP image encoder. We measure the runtime in each process: encoding, CLIP K-NN, and the updated retrieval. We can formulate each process as $\mathbf{\phi}(q)$, $\psi(\mathbf{u}, K, \mathcal{V}_2)$, and $\tilde{\psi}(\mathbf{u}, K, \mathcal{V}_2)$ respectively. We set $\hat{K} = | \mathcal{V}_2|$, the most time consuming settings. We use PyTorch libraries and execute all calculations in GPU memory. We use the same computational resources as \cref{sec:experiment}.

\cref{tab:inference-time} shows the results. Our updated retrieval $\tilde{\psi}$ takes only $3$ times overhead compared to the simple K-NN retrieval $\psi$. The encoding takes much longer than the retrieval, so the difference in the retrieval runtime can be ignored in our retrieval system. We discuss the retrieval runtime theoretically in the supplementary material.

\begin{table}[tb]
    \centering
    \caption{Average runtime of each process per query (ms). }
    \begin{tabular}{@{}llll@{}}\toprule
       Dataset & Encoding $(\mathbf{\phi}(q))$ & K-NN retrieval $(\psi(\mathbf{u}, K, \mathcal{V}_2))$ & Updated retrieval $(\tilde{\psi}(\mathbf{u}, K, \mathcal{V}_2))$     \\ \midrule
       CUB200 & $10.2$\fontsize{7}{7}$\pm0.6$  &  $0.0874$\fontsize{7}{7}$\pm0.0006$ & $0.271$\fontsize{7}{7}$\pm0.003$\\ 
       Cars-196 & $10.3$\fontsize{7}{7}$\pm 0.2$  &   $0.0875$\fontsize{7}{7}$\pm 0.0019$ &$0.272$\fontsize{7}{7}$\pm0.006$\\ \bottomrule
    \end{tabular}
    \label{tab:inference-time}
\end{table}

\section{Conclusion}
We proposed a CLIP-based interactive image retrieval system to overcome the shortcomings of metric learning. Our retrieval system receives binary feedback from each user, updates the retrieval algorithm, and returns images the user prefers. Our retrieval system adapts to any user preference and works well without training an image encoder. This paper revisited relevance feedback and integrated it with CLIP, suggesting a powerful baseline for interactive image retrieval. We believe our paper will throw a spotlight on relevance feedback again.

\bibliographystyle{splncs04}
\bibliography{main}

\begin{thebibliography}{10}
\providecommand{\url}[1]{\texttt{#1}}
\providecommand{\urlprefix}{URL }
\providecommand{\doi}[1]{https://doi.org/#1}

\bibitem{relevance-feedback-medical}
Ahmed, A.: Implementing relevance feedback for content-based medical image retrieval. IEEE Access  (2020)

\bibitem{ComposeAE}
Anwaar, M.U., Labintcev, E., Kleinsteuber, M.: Compositional learning of image-text query for image retrieval. In: WACV (2021)

\bibitem{compose-ae}
Anwaar, M.U., Labintcev, E., Kleinsteuber, M.: Compositional learning of image-text query for image retrieval. In: WACV (2021)

\bibitem{chang1971evaluation}
Chang, Y., Cirillo, C., Razon, J.: Evaluation of feedback retrieval using modified freezing, residual collection and test and control groups. The SMART retrieval system-experiments in automatic document processing  (1971)

\bibitem{beyond-personal-reidentification}
Chen, W., Chen, X., Zhang, J., Huang, K.: Beyond triplet loss: A deep quadruplet network for person re-identification. In: CVPR (2017)

\bibitem{fashion200k}
Han, X., Wu, Z., Huang, P.X., Zhang, X., Zhu, M., Li, Y., Zhao, Y., Davis, L.S.: Automatic spatially-aware fashion concept discovery. In: ICCV (2017)

\bibitem{relevance-feedback-revisited}
Harman, D.: Relevance feedback revisited. In: ACM SIGIR (1992)

\bibitem{triplet-deep}
Hoffer, E., Ailon, N.: Deep metric learning using triplet network (2014)

\bibitem{relevance-feedback-evaluation-survey}
Hull, D.: Using statistical testing in the evaluation of retrieval experiments. In: ACM SIGIR (1993)

\bibitem{MITStates}
Isola, P., Lim, J.J., Adelson, E.H.: Discovering states and transformations in image collections. In: CVPR (2015)

\bibitem{proxy-anchor}
Kim, S., Kim, D., Cho, M., Kwak, S.: Proxy anchor loss for deep metric learning. In: CVPR (2020)

\bibitem{coco-panoptic}
Kirillov, A., Lin, T.Y., Caesar, H., Girshick, R., Doll{\'a'}r, P.: Microsoft coco: Panoptic segmentation challenge (2017)

\bibitem{cars196}
Krause, J., Stark, M., Deng, J., Fei-Fei, L.: 3d object representations for fine-grained categorization. In: ICCVW (2013)

\bibitem{hist}
Lim, J., Yun, S., Park, S., Choi, J.Y.: Hypergraph-induced semantic tuplet loss for deep metric learning. In: CVPR (2022)

\bibitem{coco}
Lin, T.Y., Maire, M., Belongie, S., Hays, J., Perona, P., Ramanan, D., Doll{\'a}r, P., Zitnick, C.L.: Microsoft coco: Common objects in context. In: ECCV (2014)

\bibitem{Magface}
Meng, Q., Zhao, S., Huang, Z., Zhou, F.: Magface: A universal representation for face recognition and quality assessment. In: CVPR (2021)

\bibitem{clipcap}
Mokady, R., Hertz, A., Bermano, A.H.: Clipcap: Clip prefix for image captioning. arXiv preprint arXiv:2111.09734  (2021)

\bibitem{proxy-nca}
Movshovitz-Attias, Y., Toshev, A., Leung, T.K., Ioffe, S., Singh, S.: No fuss distance metric learning using proxies. In: ICCV (2017)

\bibitem{metric-reality-check}
Musgrave, K., Belongie, S.J., Lim, S.N.: A metric learning reality check. In: ECCV (2020)

\bibitem{revisiting-knn}
Nakata, K., Ng, Y., Miyashita, D., Maki, A., Lin, Y.C., Deguchi, J.: Revisiting a knn-based image classification system with high-capacity storage. In: ECCV (2022)

\bibitem{CLIP}
Radford, A., Kim, J.W., Hallacy, C., Ramesh, A., Goh, G., Agarwal, S., Sastry, G., Askell, A., Mishkin, P., Clark, J., Krueger, G., Sutskever, I.: Learning transferable visual models from natural language supervision. In: ICML (2021)

\bibitem{relevance-feedbacl-powerful-tool}
Rui, Y., Huang, T., Ortega, M., Mehrotra, S.: Relevance feedback: a power tool for interactive content-based image retrieval. IEEE TCSVT  (1998)

\bibitem{sb-zsir}
Sain, A., Bhunia, A.K., Chowdhury, P.N., Koley, S., Xiang, T., Song, Y.Z.: {CLIP for All Things Zero-Shot Sketch-Based Image Retrieval, Fine-Grained or Not}. In: CVPR (2023)

\bibitem{pic2word}
Saito, K., Sohn, K., Zhang, X., Li, C.L., Lee, C.Y., Saenko, K., Pfister, T.: Pic2word: Mapping pictures to words for zero-shot composed image retrieval. CVPR  (2023)

\bibitem{genecis}
Vaze, S., Carion, N., Misra, I.: Genecis: A benchmark for general conditional image similarity. In: CVPR (2023)

\bibitem{clipasso}
Vinker, Y., Pajouheshgar, E., Bo, J.Y., Bachmann, R.C., Bermano, A.H., Cohen-Or, D., Zamir, A., Shamir, A.: Clipasso: Semantically-aware object sketching. ACM Trans. Graph.  (2022)

\bibitem{tirg}
Vo, N., Jiang, L., Sun, C., Murphy, K., Li, L.J., Fei-Fei, L., Hays, J.: Composing text and image for image retrieval - an empirical odyssey. In: CVPR (2019)

\bibitem{multi-similarity}
Wang, X., Han, X., Huang, W., Dong, D., Scott, M.R.: Multi-similarity loss with general pair weighting for deep metric learning. In: CVPR (2019)

\bibitem{triplet}
Weinberger, K.Q., Blitzer, J., Saul, L.: Distance metric learning for large margin nearest neighbor classification. In: NIPS (2005)

\bibitem{CUB200}
Welinder, P., Branson, S., Mita, T., Wah, C., Schroff, F., Belongie, S., Perona, P.: Caltech-ucsd birds 200  (2010)

\bibitem{few-show-feature-map}
Wertheimer, D., Tang, L., Hariharan, B.: Few-shot classification with feature map reconstruction networks. In: CVPR (2021)

\bibitem{gldv2}
Weyand, T., Araujo, A., Cao, B., Sim, J.: {Google Landmarks Dataset v2 - A Large-Scale Benchmark for Instance-Level Recognition and Retrieval}. In: CVPR (2020)

\bibitem{multilevel-relevance-feedback}
Wu, H., Lu, H., Ma, S.: Willhunter: interactive image retrieval with multilevel relevance. In: ICPR (2004)

\end{thebibliography}

\clearpage

\newcommand\beginsupplement{%
        \setcounter{table}{0}
        \renewcommand{\thetable}{\Alph{table}}%
        \setcounter{figure}{0}
        \renewcommand{\thefigure}{\Alph{figure}}%
        \setcounter{section}{0}
        \renewcommand{\thesection}{\Alph{section}}
        \setcounter{chapter}{0}
        \renewcommand{\thechapter}{\Alph{chapter}
        }
     }

\beginsupplement

\chapter*{Appendix}

\section{Implementation Details}

To increase the reproducibility of our experiments and make them more understandable, we provide additional information and explanations about our experimental conditions that we omit due to space limitations.

\subsection{Category-based image retrieval}

\begin{figure}[tb]
    \centering
    \includegraphics[width=0.7\linewidth]{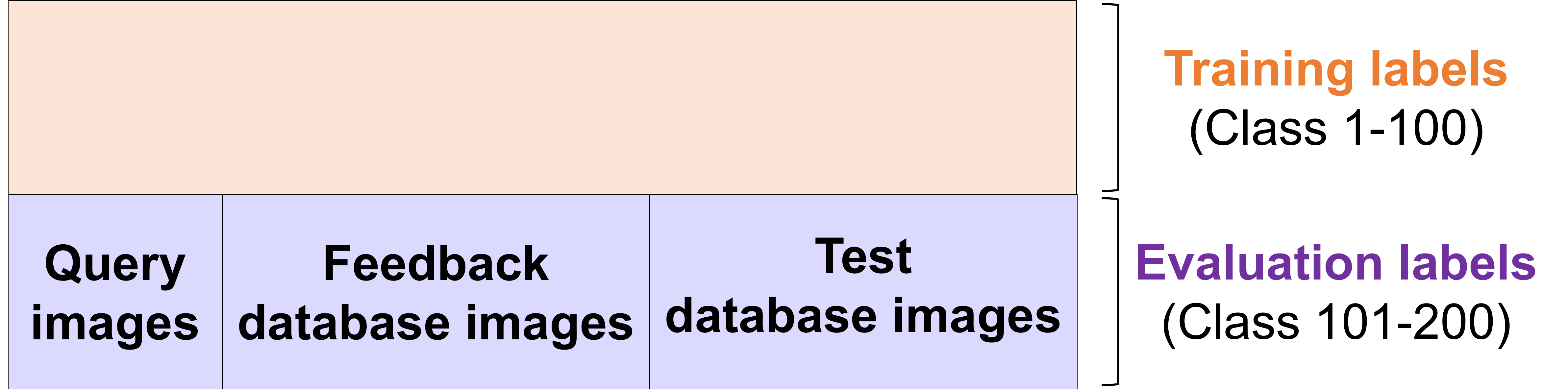}
    \caption{Dataset usage in case of CUB-200-2011.}
    \label{fig:sup-dataset-usage-cub200}
\end{figure}

For the implementation of metric-learning-based methods, when training an encoder with metric learning, we follow the common practice of metric learning and treat the former half of the labels as the training labels and the latter half as the evaluation labels. \cref{fig:sup-dataset-usage-cub200} illustrates the dataset usage in our category-based image retrieval experiment in the case of CUB-200-2011. CUB-200-2011 has $200$ labels in total. Therefore, we use the first $100$ labels for training and the second $100$ labels for evaluation. We treat images of the second $100$ labels as an evaluation dataset, and execute the dataset splitting to obtain three subsets: query images, feedback database images, and test database images. 

Since feedback database images are reserved for applying relevance feedback as in our proposed approach, when evaluating metric-learning-based image retrieval approaches, only query images and test database images have been used. 

\subsection{One-label-based image retrieval}\label{sup:subsec:implementation-detal-one-label-based}
For MIT-States and Fashion200k datasets, some images have been excluded in our evaluation because their captions, i.e., the set of labels, are not qualified. The details are as below. 

First, we exclude rare captions that are not shared by enough images. Specifically, if a caption occurs less than 5 times in the evaluation dataset, where 5 is the minimal number of samples for doing a 1:2:2 splitting introduced in Sec. 4, the few images with that caption are not included in our experiments. For example, there are only two images with the "molten orange" caption in the dataset, so the two images are excluded in our experiments. 

Second, when using MIT-States, we exclude samples with the adjective label ``adj.'' This is the same implementation as the prior conditioned image retrieval studies.

\subsection{Conditioned image retrieval}

To have MIT-States dataset for conditioned image retrieval evaluation, we first remove the images with unqualified captions as described in \cref{sup:subsec:implementation-detal-one-label-based}. Additionally, when a query image has a noun label $l_{\mathrm{noun}}$ and an adjective label $l_{\mathrm{adj}}$, we choose a different adjective label $l_{\mathrm{adj}}'$ in the condition that there is at least one sample that has both $l_{\mathrm{adj}}'$ and $l_{\mathrm{noun}}$. For example, no sample has both ``unripe'' and ``mountain'' in the dataset, so we do not choose $l_{\mathrm{adj}}'=\text{``unripe''}$ when we handle the query image which has $l_{\mathrm{noun}}=\text{``mountain.''}$

\section{Dataset Details}
\begin{table}[tb]
    \centering
    \caption{Dataset details of category-based image retrieval}
    \label{tab:category-dataset-details}
    \begin{tabular}{@{}p{10em}p{6em}p{6em}@{}} \toprule
       Dataset  &  \# labels & \# images  \\ \midrule
        CUB-200-2011 & $100$ & $5924$ \\
        Cars-196     & $98$  & $8131$ \\ \bottomrule
    \end{tabular}
\end{table}

\begin{table}[tb]
    \centering
    \caption{Dataset details of one-label-based image retrieval}
    \label{tab:one-label-based-dataset-details}
    \begin{tabular}{@{}p{10em}p{6em}p{6em}p{10em}p{6em}@{}} \toprule
       Dataset  &  \# labels & \# images & types of labels & \# queries  \\ \midrule
        Fashion200k & $1258$ & $10609$ & adjective, noun & $9722$\fontsize{7}{7}$\pm5$ \\
        MIT-States     & $146$  & $10443$ & adjective, noun & $4178$\fontsize{7}{7}$\pm0$ \\ 
        COCO         &   $133$     &  $5000$  & noun & $6967$\fontsize{7}{7}$\pm196$ \\\bottomrule
    \end{tabular}
\end{table}

\begin{table}[tb]
    \centering
    \caption{Dataset details of conditioned image retrieval}
    \label{tab:conditioned-dataset-details}
    \begin{tabular}{@{}p{10em}p{6em}p{6em}p{10em}p{6em}@{}} \toprule
       Dataset  &  \# labels & \# images & types of labels & \# queries  \\ \midrule
        MIT-States     & $146$  & $10443$ & adjective, noun & $15368$\fontsize{7}{7}$\pm2$ \\\bottomrule
    \end{tabular}
\end{table}

We describe the details of the evaluation datasets that we use in our experiments in \cref{tab:category-dataset-details,tab:one-label-based-dataset-details,tab:conditioned-dataset-details}. We also attach information about the number of queries in \cref{tab:one-label-based-dataset-details,tab:conditioned-dataset-details}. We change the splitting way ten times and calculate the average and standard deviation of the number of queries because the number of queries depends on the number of query images' labels. 

It is not a mistake that the number of queries in MIT-States differs between one-label-based image retrieval and conditioned image retrieval. As described in our main paper, the number of queries for one query image differs in these two settings. In one-label-based image retrieval, when a query image has $L$ labels, we execute $L$ queries for all labels of the query image. In conditioned image retrieval, when a query image has $l_{\mathrm{adj}}$, we execute queries for all possible different adjective labels $l_{\mathrm{adj}}' \in \mathcal{L}_{\mathrm{adj}} \backslash \{ l_{\mathrm{adj}} \}$.

\section{Examples Cases}
\subsection{One-label-based image retrieval}
\begin{figure}[tb]
    \centering
    \begin{minipage}[c]{\linewidth}
        \centering
        \includegraphics[width=0.8\linewidth]{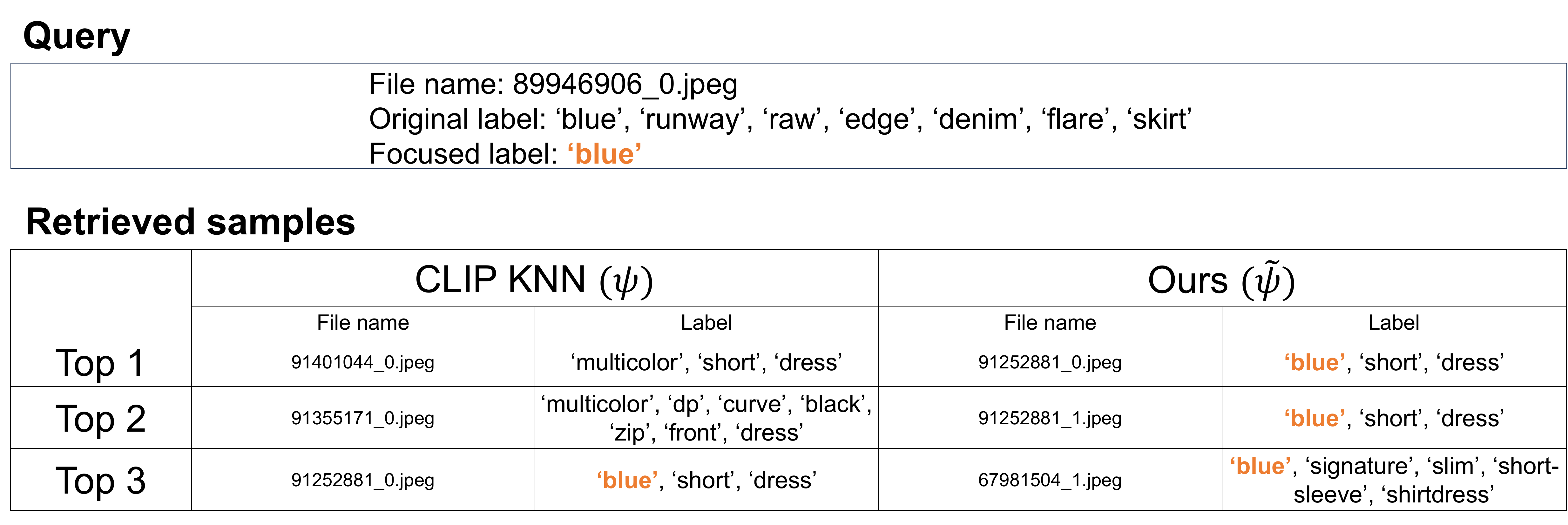}
        \caption*{Fashion200k}
        \label{sup:fig:additional-examples-fashion200k}
    \end{minipage}
    \begin{minipage}[c]{\linewidth}
        \centering
        \includegraphics[width=0.8\linewidth]{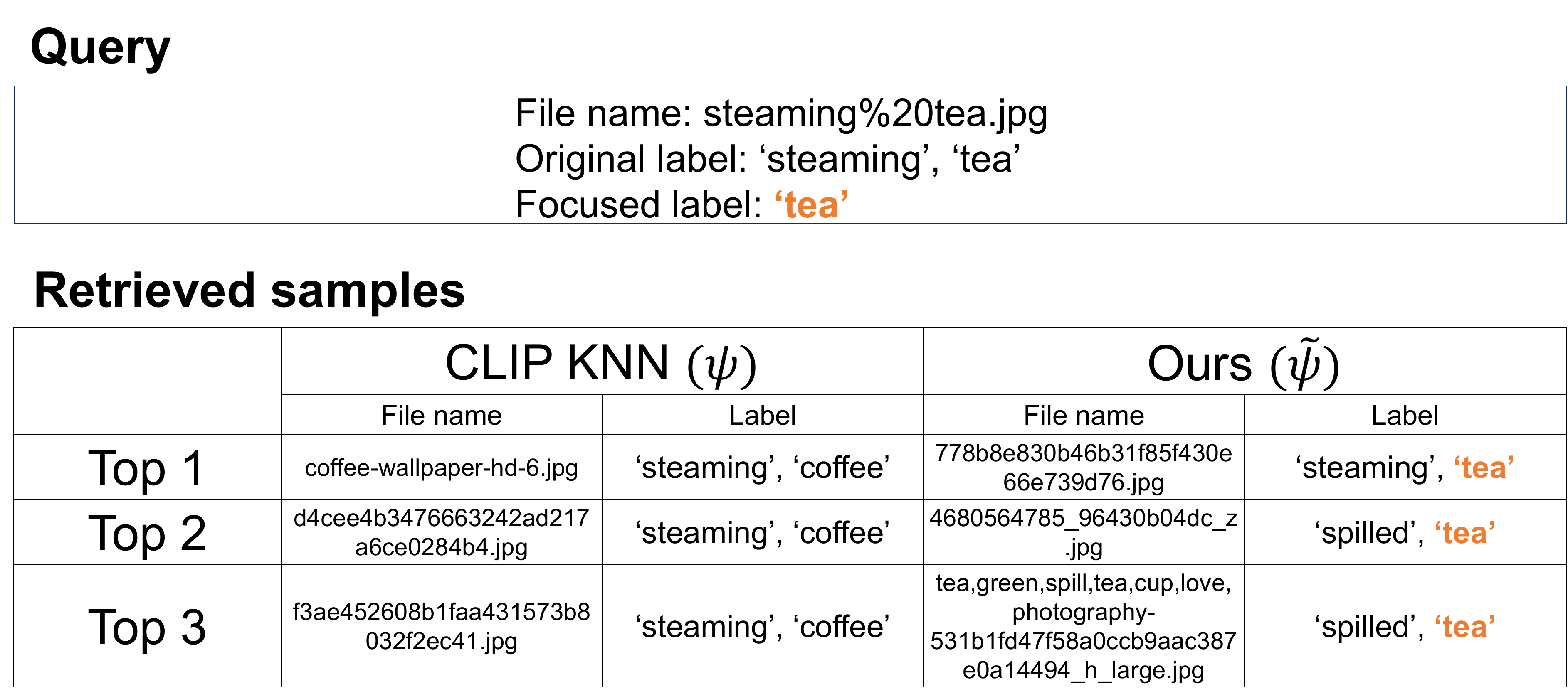}
        \caption*{MIT-States}
        \label{sup:fig:additional-examples-mit}
    \end{minipage}
    \caption{Examples of one-label-based image retrieval.}
    \label{sup:fig:additional-examples}
\end{figure}

We provide additional examples of one-label-based image retrieval in \cref{sup:fig:additional-examples}. In the case of Fashion200k, we simulate a user who searches for images of a blue object by providing the query image to our retrieval system. Our retrieval system successfully learns the user's preference from the feedback, and returns images with ``blue'' labels more accurately than image retrieval without relevance feedback. In the case of MIT-States, we simulate a user who searches for tea images by providing the query image of streaming tea to our retrieval system. Our retrieval system successfully learns the user's preference from the feedback, and returns tea images more accurately.

\subsection{Conditioned image retrieval}

\begin{figure}[tb]
    \centering
    \includegraphics[width=0.8\linewidth]{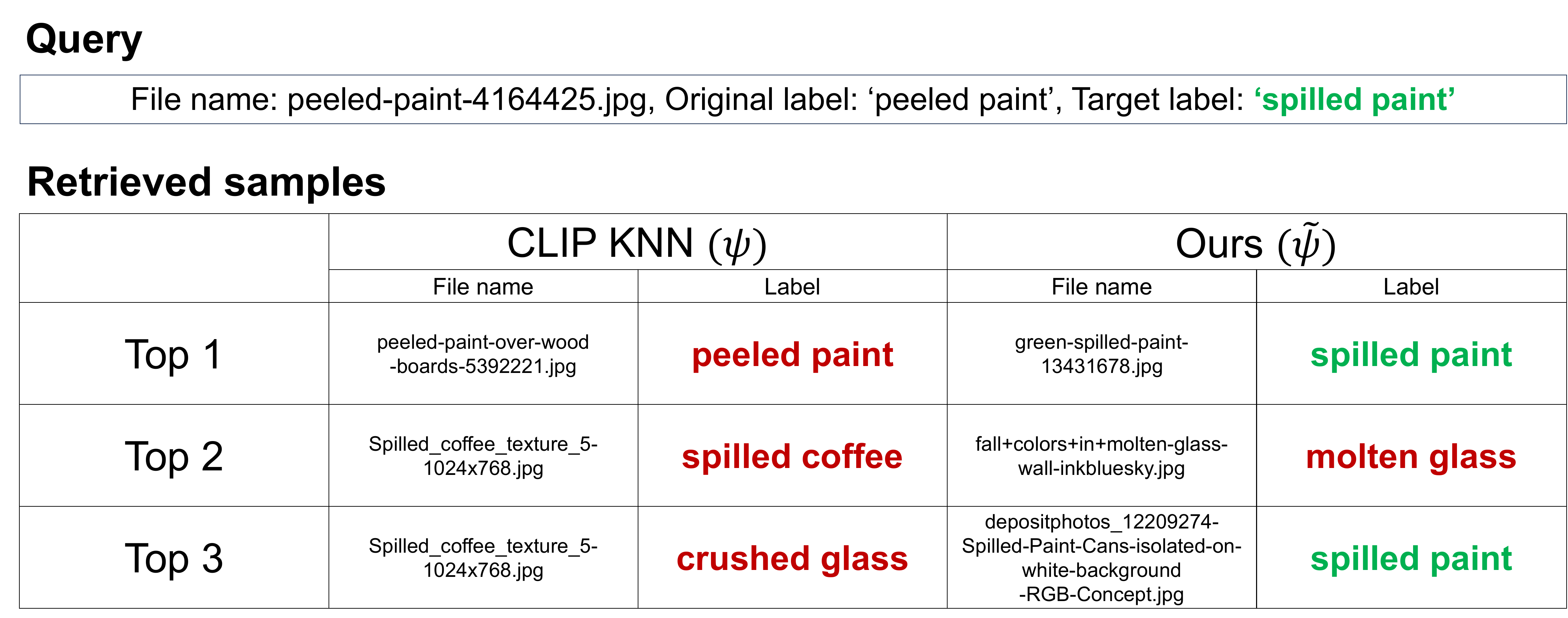}
    \caption{An example of conditioned image retrieval.}
    \label{fig:conditioned-sample}
\end{figure}

We demonstrate an example trial of our retrieval system (CLIP ViT-B/16) in \cref{fig:conditioned-sample}. In the first case, we choose an adjective label ``spilled,'' simulating a user who provides the peeled paint image searching for spilled paint images. With relevance feedback, our retrieval system can return images with the labels ``spilled'' and ``paint'' more accurately.

% \section{Feedback Size Suitability}
% Here, we would also like to explain our decision on performing setup of $M=50$ in each experiment. From our conversations with industrial engineers, $M=50$ is found to be within optimal values for many industrial applications such as defects analysis. It is because $M=50$ is large enough to provide useful macro analysis in a single round as engineers need to check the search results anyway, and small enough for feedbacks creation. We would like to point out that by defaulting feedback label to ``False'', only $8.62$, $5.99$ and $3.26$ operations in average are needed to flip positive samples to ``True'' for $M=50, 25, 10$ respectively in our experiment (CUB-200-2011). Furthermore, the binary nature of the feedback allows more efficient feedback methods, such as signals from user's brain activity.

\section{Theoretical and Additional Runtime Analysis}

\begin{table}[tb]
    \centering
    \caption{Comparison of $\mathrm{Recall@1}$ and retrieval runtime (ms) among various $\hat{K}$. $|\mathcal{V}_2|$ is around $2000$ in CUB-200-2011, and around $3000$ in Cars-196.}
    \begin{tabular}{@{}p{10em}p{5em}p{5em}p{12em}p{6em}@{}}\toprule
         Dataset & Method & $\hat{K}$ & Retrieval runtime (ms) &  $\mathrm{Recall@1}$ \\ \midrule
          CUB-200-2011    & $\psi$ & - & $0.0874$\fontsize{7}{7}$\pm0.0006$ & $51.9$\fontsize{7}{7}$\pm1.0$ \\ \cmidrule(){2-5}
              & $\tilde{\psi}$ & $30$ & $0.232$\fontsize{7}{7}$\pm0.006$ & $69.3$\fontsize{7}{7}$\pm1.6$ \\
              &  & $100$ & $0.253$\fontsize{7}{7}$\pm0.005$ & $70.8$\fontsize{7}{7}$\pm1.4$ \\
              &  & $|\mathcal{V}_2|$ & $0.272$\fontsize{7}{7}$\pm0.006$ & $70.9$\fontsize{7}{7}$\pm1.4$ \\ \midrule
          Cars-196    & $\psi$ & - & $0.0875$\fontsize{7}{7}$\pm0.0019$ & $72.9$\fontsize{7}{7}$\pm0.9$ \\ \cmidrule(){2-5}
              & $\tilde{\psi}$ & $30$ & $0.255$\fontsize{7}{7}$\pm0.004$ & $87.2$\fontsize{7}{7}$\pm1.0$ \\
              &  & $100$ & $0.266$\fontsize{7}{7}$\pm0.005$ & $87.7$\fontsize{7}{7}$\pm0.9$ \\
              &  & $|\mathcal{V}_2|$ & $0.272$\fontsize{7}{7}$\pm0.006$ & $87.7$\fontsize{7}{7}$\pm0.9$ \\ \bottomrule
    \end{tabular}
    \label{tab:retrieval_time_recall1_hatK}
\end{table}

% We explain theoretically why the updated retrieval $(\tilde{\psi})$ is slower than simple KNN $(\psi)$ as shown in \cref{subsec:retrieval-time}. 
We provide a theoretical explanation of the retrieval runtime difference between K-NN $(\psi(\mathbf{u}, K, \mathcal{V}_2))$ and the updated retrieval $(\tilde{\psi}(\mathbf{u}, K, \mathcal{V}_2)))$.
Simple K-NN calculates cosine similarities $|\mathcal{V}_2|$ times. In contrast, the updated retrieval executes cosine similarities $\hat{K}M$ times.
The updated retrieval calculates cosine similarities $\hat{K}M+|\mathcal{V}_2|$ times. Therefore, the updated retrieval is $1 + \frac{\hat{K}M}{|\mathcal{V}_2|}$ times slower than the simple K-NN. In our experiment, we set $M=50$ and $\hat{K}=|\mathcal{V}_2|$, so the updated retrieval runtime could be $51$ times longer than the simple K-NN retrieval runtime.
In our experiment, we execute retrieval in GPU memory. GPUs are good at simple parallel computation, and since the number of data we are dealing with in this paper can be fitted into GPU memory, there is not much difference in the retrieval runtime.

To explore retrieval runtime further, we try smaller $\hat{K}$ and observe retrieval accuracy and speed. We set $\hat{K}=30,100,|\mathcal{V}_2|$ and compare retrieval runtime and $\mathrm{Recall@1}$. \cref{tab:retrieval_time_recall1_hatK} shows the results. Larger $\hat{K}$ leads to longer retrieval runtime, this can be expected from our theoretical explanation. Additionally, setting smaller $\hat{K}$ slightly lower $\mathrm{Recall@1}$.

\section{Potential Limitations and Future Works}

\subsection{Potential Limitations}
Our retrieval may underperform in certain scenarios. Our retrieval system relies on CLIP's zero-shot transferability, but CLIP does not work well in some datasets, such as specialized industrial datasets. Exploring what image encoders our retrieval system should use instead of CLIP to adapt such scenarios is an important future work.

Another potential drawback of our retrieval system is that it requires binary feedback from the user. The necessity of binary feedback could worsen user experience, especially when users want to find information quickly and easily. To reduce user efforts, we should consider more effective ways to convey each user's preference than binary feedback.

\subsection{Future Works}
This paper only focuses on one-time binary relevance feedback. In future works, we aim to consider more complex interactive image retrieval tasks by expanding our retrieval system to incorporate multi-level and multi-turn relevance feedback. 

Additionally, experiments with actual user feedback are necessary for a more accurate evaluation of our retrieval system. In this study, we prepare three experimental settings and automatically generate feedback to simulate users' feedback. However, actual users may have various preferences that cannot be fully replicated by current methods. By having actual users provide feedback on returned samples, we can evaluate the accuracy of our retrieval system in more practical situations.

\end{document}